\documentclass[10pt,twocolumn,letterpaper]{article}

\usepackage{cvpr}      % To produce the REVIEW version
\usepackage{times}
\usepackage[normalem]{ulem}
\usepackage{url}
\usepackage{graphicx}
\usepackage[most]{tcolorbox}
\usepackage{mdframed}
\usepackage[table,xcdraw,usenames,dvipsnames]{xcolor}
\usepackage{booktabs}
\usepackage{multirow}
\usepackage{makecell}
\usepackage{colortbl}
\usepackage{siunitx}
\usepackage{amssymb}
\usepackage{minted}
\usepackage[ruled,vlined,noend]{algorithm2e}

\usepackage{pifont}

\newcolumntype{b}{>{\columncolor{blue!10}}c}
\SetKwComment{Comment}{$\triangleright$ }{}

\SetCommentSty{mycommfont}

\tcbset{
  custombox/.style 2 args={
    colback=gray!8!white,
    colframe=white,
    boxrule=0.5mm,
    arc=2mm,
    boxsep=3mm,
    drop shadow={black!20!white},
    enhanced,
    overlay={
      \node[
        fill=#1,       % 第1个参数：颜色
        text=white,
        rounded corners=0.8mm,
        font=\bfseries\scriptsize,
        inner sep=1mm
      ] at (frame.north west) [xshift=9mm, yshift=-0.8mm] {#2}; % 第2个参数：标签文字
    }
  }
}

\definecolor{cvprblue}{rgb}{0.21,0.49,0.74}
\usepackage[pagebackref,breaklinks,colorlinks,allcolors=cvprblue]{hyperref}

\def\paperID{11206} % *** Enter the Paper ID here
\def\confName{CVPR}
\def\confYear{2026}

\title{Graph2Eval: Automatic Multimodal Task Generation for Agents \\ via Knowledge Graphs}

\author{
Yurun Chen\textsuperscript{1} \quad 
Xueyu Hu\textsuperscript{1} \quad  
Yuhan Liu\textsuperscript{2} \quad  
Ziqi Wang\textsuperscript{1} \quad 
Zeyi Liao\textsuperscript{3} \quad  
Lin Chen\textsuperscript{4}  \\
Feng Wei\textsuperscript{4} \quad 
Yuxi qian\textsuperscript{4} \quad
Bo Zheng\textsuperscript{4} \quad
Keting Yin\textsuperscript{1,}\thanks{Corresponding Author} \quad
Shengyu Zhang\textsuperscript{1,}\footnotemark[1]\\
\textsuperscript{1}Zhejiang University \quad
\textsuperscript{2}Xiameng University \quad
\textsuperscript{3}The Ohio State University \quad
\textsuperscript{4}Ant Group \\
Project Page: \url{https://github.com/YurunChen/Graph2Eval}
}

\begin{document}

\maketitle

\begin{abstract}

As multimodal LLM-driven agents advance in autonomy and generalization, traditional static datasets face inherent scalability limitations and are insufficient for fully assessing their capabilities in increasingly complex and diverse tasks. Existing studies have attempted to generate agent tasks using LLMs, but due to the inherent hallucinations of LLMs and the lack of internal data relationship modeling, these tasks often exhibit semantic inconsistencies and solvability issues. To address these challenges, we introduce \textsc{Graph2Eval}, a knowledge-graph-driven framework for automated, scalable, and semantically grounded agent task generation. At its core, \textsc{Graph2Eval} leverages a knowledge graph built from heterogeneous external data sources as a structured task space, generating multimodal agent tasks through subgraph sampling and task construction guided by task templates and meta-path strategies. To further ensure task reliability, a multi-stage filtering pipeline based on node reachability analysis, LLM scoring, and similarity analysis ensures the diversity and solvability of the generated tasks. By unifying both RAG Agent and Web Agent scenarios, \textsc{Graph2Eval} enables efficient generation of multimodal document understanding tasks and multi-step web interaction tasks. We instantiate the framework with \textsc{Graph2Eval-Bench}, a curated dataset of 1,319 tasks spanning document understanding and web interaction scenarios. Extensive experiments show that, on average, \textsc{Graph2Eval} improves task semantic consistency by 20\% and solvability by 17\% over baselines, while \textsc{Graph2Eval-Bench} effectively distinguishes agent performance, offering a new perspective on agent evaluation.
\end{abstract}

\vspace{-0.5cm}
\section{Introduction}
A static and narrow evaluation framework risks mistaking surface-level familiarity for true cognitive ability. In a simplified analogy, repeatedly presenting the same fixed set of exercises may allow a person to memorize answers and appear to perform perfectly, while their true adaptability remains untested. The same concern arises for multimodal LLM-based agents: While large-scale pretraining on foundation models and domain-specific finetuning have substantially advanced agent performance~\citep{ liu2025infiguiagentmultimodalgeneralistgui, liu2025infiguir1advancingmultimodalgui, choudhury2024betterteacherllmagents, chen2025harmonyguard, hu2025agents}, existing static datasets fail to disentangle whether an agent’s task success reflects genuine generalization or mere retrieval of memorized knowledge, thereby undermining the assessment of true competence. Therefore, success on static datasets does not imply that an agent can generalize or remain reliable in real-world scenarios.

\begin{figure*}[t]
\centering
\includegraphics[width=0.95\linewidth]{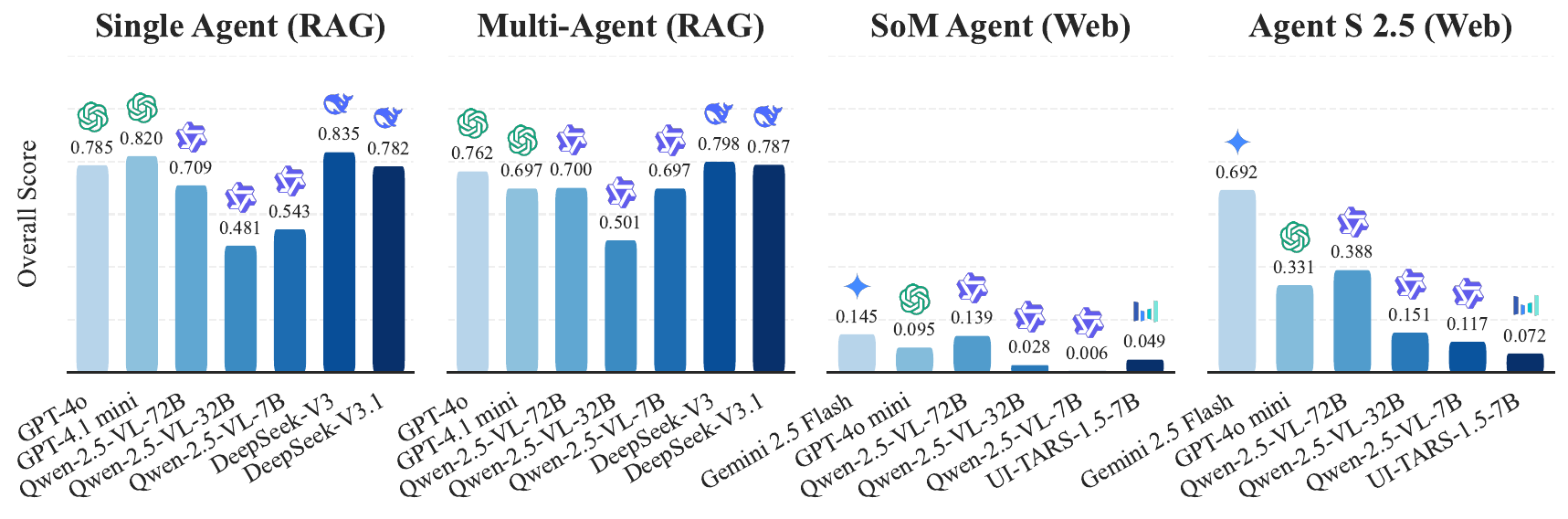}
\caption{\textsc{Graph2Eval-Bench} supports evaluating agent performance across different foundation models.}
\label{fig:abs_performance}
\end{figure*}

Rapidly expanding and updating datasets could reduce evaluation bias from agents’ reliance on internal training knowledge, yet most current datasets offer limited support for such dynamic capabilities. Construction of static datasets remains heavily dependent on manual annotation or reuse of prior resources~\citep{mialon2023gaiabenchmarkgeneralai, deng2023mind2webgeneralistagentweb, liu2023agentbench, yan2025mcpworldunifiedbenchmarkingtestbed, chen2025shieldagentshieldingagentsverifiable, xiang2025guardagentsafeguardllmagents}, and online environment datasets~\citep{zhou2023webarena, OSWorld, liao2025redteamcuarealisticadversarialtesting, evtimov2025waspbenchmarkingwebagent, tur2025safearenaevaluatingsafetyautonomous, levy2025stwebagentbenchbenchmarkevaluatingsafety, boisvert2025doomarenaframeworktestingai} also demand substantial human effort to expand tasks, such as building controlled environments and designing content-specific challenges. These constraints limit task diversity and alignment with dynamic environments, motivating the development of automated agent task generation methods to alleviate such labor-intensive bottlenecks~\citep{shi2025taskcraftautomatedgenerationagentic, li2025agenthospitalsimulacrumhospital, zhang2025agentracerinducingfailurellm,su2025learnbyinteractdatacentricframeworkselfadaptive}. However, these methods face two major limitations:
\textbf{(I) Absence of explicit entity-relation modeling}: Current methods generally rely on preprocessed text and image data, which are directly analyzed by LLMs for task generation, without explicitly modeling the semantic structure among entities. As a result, the generated tasks often suffer from semantic inconsistency and limited solvability, and fail to capture the complex interdependencies required for higher-order reasoning and cognitive evaluation.
\textbf{(II) Limited adaptation to dynamic environments}: Current methods for generating web interaction tasks rely on static data with predefined page relationships or on LLMs analyzing simplified environments, and fail to explore effective modeling of content and page relationships in real-world websites. As a result, the generated tasks are difficult to transfer to dynamic web scenarios, preventing reliable evaluation of agents’ generalization and performance in real-world environments.

To address these gaps, we propose \textbf{\textsc{Graph2Eval}}, a Knowledge Graph (KG)-based framework for automatic agent task generation that ensures semantic consistency and task solvability. In this framework, the KG serves as the \textit{task space} for task generation, enabling the creation of document understanding and web interaction tasks. Specifically, \textsc{Graph2Eval} employs a unified graph abstraction to encode entities, relations, and interactions from textual and web data as nodes and edges representing both semantic and interactive elements. Building on this graph, \textsc{Graph2Eval} defines task structures using task templates and meta-paths, which specify the types and order of nodes in a task and guide structured task generation. Subgraph sampling strategies are then applied to extract the required nodes and edges. Subsequently, LLMs integrate the sampled subgraph structures with contextual information via context engineering, generating diverse and well-formed task instances.

We implemented a prototype of \textsc{Graph2Eval} and constructed \textsc{Graph2Eval-Bench}, a dataset containing 1,319 diverse tasks, including 1,002 document understanding tasks and 317 web interaction tasks, which is designed to evaluate both Retrieval-Augmented Generation~(RAG) Agents and Web Agents. Extensive experiments demonstrate that, compared to a KG-free baseline, tasks generated by \textsc{Graph2Eval} achieve semantic consistency and solvability improvements of 20\% and 17\% on average. In addition, \textsc{Graph2Eval} achieves high efficiency, requiring on average 34.87 seconds per document understanding task and 95.51 seconds per web interaction task. Furthermore, \textsc{Graph2Eval-Bench} effectively distinguishes agent performance across different LLM configurations (Figure~\ref{fig:abs_performance}). Our contributions are summarized as follows:

\begin{itemize}[leftmargin=*]
    \item We propose a novel perspective for agent task generation by treating KGs constructed from multi-source data as a latent task space, aiming to improve semantic consistency and solvability in synthetic agent tasks.
    \item We introduce \textsc{Graph2Eval}, a KG–based framework that exploits semantic relations within data for automatic agent task generation, providing a unified pipeline for rapid dataset creation in a reproducible manner.
    \item We implement a full prototype of \textsc{Graph2Eval} and construct \textsc{Graph2Eval-Bench}, a dataset of 1,319 agent tasks designed for evaluating both RAG Agents and Web Agents. Extensive experiments show that the framework efficiently generates diverse tasks, improves semantic consistency and solvability, and effectively differentiates performance across various agents.
\end{itemize}

% \begin{figure*}[t]
%     \centering
%     \includegraphics[width=1\textwidth]{figures/strategy4.pdf}
%     \caption{Overview of the dataset generated by \textsc{Graph2Eval}.}
%     \label{fig:strategy}
% \end{figure*}

\section{Background}
This section highlights the limitations of existing datasets in terms of scalability, semantic consistency, and support for dynamic scenario tasks for agents.

\vspace{-0.3cm}
\paragraph{Human-Annotated Data.} 
Most existing evaluation benchmarks rely on human annotation. 
In the LLM era, datasets~\cite{wang2024mmlu,rein2024gpqa, liu2024mmbenchmultimodalmodelallaround,du2025twinvoice} mainly focus on static task evaluation. 
As research shifts toward tool-using~\cite{yao2024tau, patil2025the, wang2026ragrouter} and web agents, new agent-oriented benchmarks have emerged, including GAIA~\citep{mialon2023gaiabenchmarkgeneralai}, MiniWob~\citep{pmlr-v70-shi17a}, MiniWoB++~\citep{liu2018reinforcementlearningwebinterfaces}, and Mind2Web~\citep{deng2023mind2webgeneralistagentweb}, most of which are manually constructed. 
Some benchmarks further incorporate realistic environments, such as OS World~\citep{xie2024osworldbenchmarkingmultimodalagents}, AndroidWorld~\citep{rawles2025androidworlddynamicbenchmarkingenvironment}, RedTeamCuA~\citep{liao2025redteamcuarealisticadversarialtesting}, WASP~\cite{evtimov2025waspbenchmarkingwebagent}, SafeArena~\citep{tur2025safearenaevaluatingsafetyautonomous}, ST-WebAgentBench~\citep{levy2025stwebagentbenchbenchmarkevaluatingsafety}, and DoomArena~\citep{boisvert2025doomarenaframeworktestingai}. 
Despite richer environments, their task specifications remain predominantly human-defined, limiting scalability.

\vspace{-0.3cm}
\paragraph{Synthetic Data.} Some approaches leverage LLMs to synthesize inputs in various ways, including seed-task-based generation~\citep{wang2023selfinstructaligninglanguagemodels, xu2023baizeopensourcechatmodel,li2024numinamath,toshniwal2024openmathinstruct2acceleratingaimath}, question rewriting~\citep{yu2024metamathbootstrapmathematicalquestions}, and self-iterative methods~\citep{zelikman2022starbootstrappingreasoningreasoning,qiao2024autoactautomaticagentlearning}. Additionally, other methods enhance instruction complexity through rule-based modifications~\citep{xu2025wizardlmempoweringlargepretrained} or extract question-answer pairs from web-pretrained corpora to construct training data~\citep{yue2024mammoth2scalinginstructionsweb}. However, these synthetic data generation techniques primarily focus on training or evaluating LLMs rather than on agent-centric tasks. TaskCraft~\citep{shi2025taskcraftautomatedgenerationagentic} automates the construction of tool-using agent tasks by first generating atomic tasks via LLM-based document traversal, and then expanding the task space through task composition. Nevertheless, this compositional approach still relies on the LLM’s judgment of inter-task dependencies and lacks an effective understanding of explicit entity relationships. Similarly, \citet{su2025learnbyinteractdatacentricframeworkselfadaptive} synthesize agent-environment interaction data from documents using LLMs, but the quality of the generated tasks is ensured only through deduplication and LLM-based alignment and evaluation. Moreover, since this method generates web agent tasks based on static documents, it lacks an understanding of tasks in actual dynamic scenarios and therefore cannot reliably assess task executability.

\section{\textsc{Graph2Eval}}
\label{sec:graph2eval}

\begin{figure*}[t]
    \centering
    \includegraphics[width=1\textwidth]{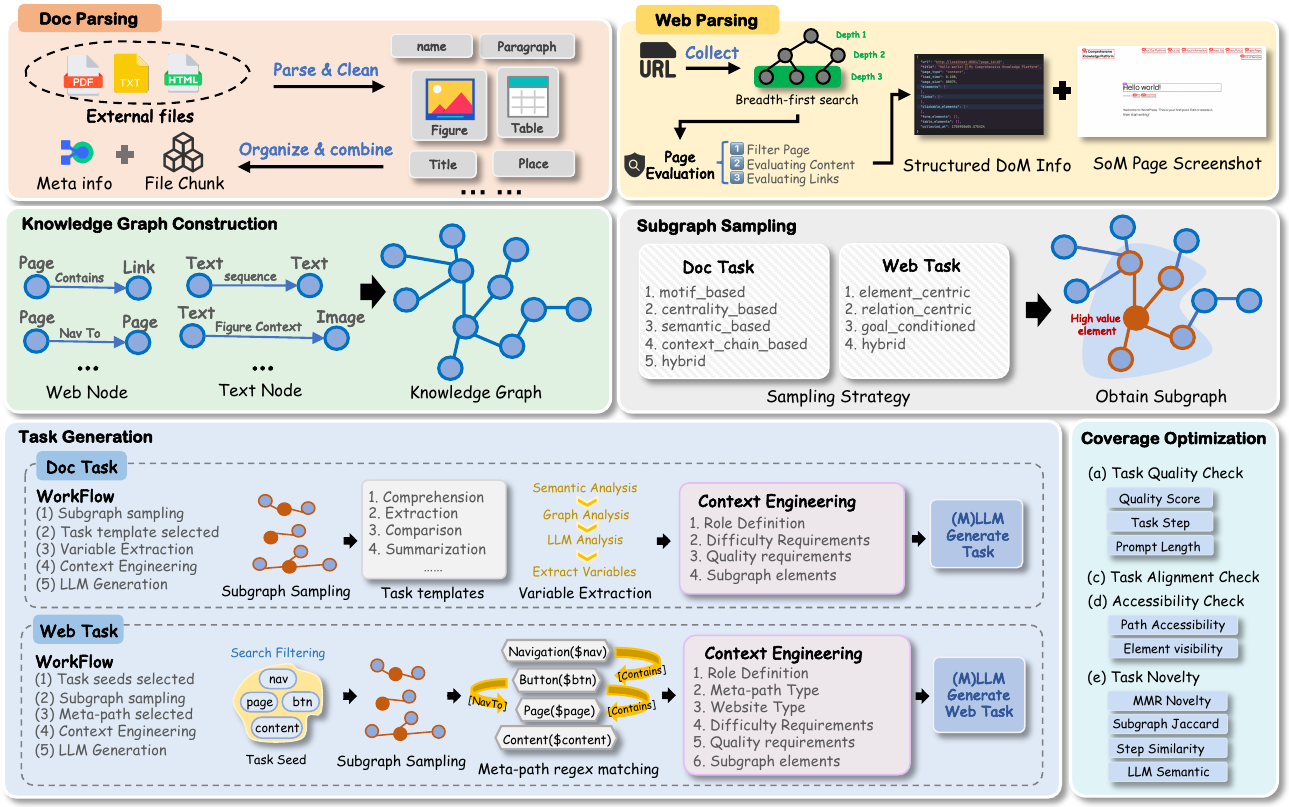}
    \caption{Workflow for dataset generation in \textsc{Graph2Eval}: (1)~Data Ingestion (\textbf{Top Left~/~Right}): parsing documents and crawling web pages to extract structured content. (2)~KG Construction (\textbf{Middle Left}): building the graph by identifying nodes and edges that encode semantic, structural, and interactive relations. (3)~Subgraph Sampling (\textbf{Middle Right}): applying scenario-specific sampling strategies for document and web tasks based on execution modes. (4)~Task Generation (\textbf{Bottom Left}): instantiating and composing tasks from sampled subgraphs, producing diverse, executable task units. (5)~Coverage Optimization (\textbf{Bottom Right}): evaluating and selecting generated tasks to ensure quality, diversity, and representativeness.}
    \label{fig:method}
\end{figure*}

\textsc{Graph2Eval} aims to provide an efficient task generation framework that addresses challenges in task semantic alignment and solvability in existing agent task generation methods. \textsc{Graph2Eval} is characterized by two key features:
\textbf{(1) Knowledge-Graph–Driven Task Space}: Tasks are generated based on a structured knowledge graph constructed from multi-source external data, capturing entities and their semantic relationships to ensure consistent and executable task generation.
\textbf{(2) Multi-Scenario Task Generation}: \textsc{Graph2Eval} extracts entities and their dependencies from documents to generate document understanding tasks for RAG Agents, and produces web interaction tasks for Web Agents that reflect realistic deployment scenarios. The dataset generation workflow comprises five stages: \textit{data ingestion $\rightarrow$ KG construction $\rightarrow$ subgraph sampling $\rightarrow$ task generation $\rightarrow$ coverage optimization}, as illustrated in Figure~\ref{fig:method}.

% \subsection{Dataset Generation}

\subsection{Data Ingestion} 
During preprocessing, \textsc{Graph2Eval} structures document content beyond plain text by preserving hierarchical semantics and layout elements such as paragraphs, tables, headings, and figure captions. This involves three key steps: \textbf{(1) Semantic Chunking}, segmenting the document into minimal semantic units mapped to graph nodes; \textbf{(2) Embedding Computation}, encoding each node with deep semantic embeddings to capture contextual dependencies; and \textbf{(3) Metadata Annotation}, enriching nodes with source and positional metadata (\textit{e.g.}, file path, title, author). This \textit{content → node → embedding + metadata} representation ensures semantic fidelity and cross-document consistency, enabling KG construction and task generation. For web data, pages are collected via automated URL crawling, extracting DoM structures and screenshots. To handle complex modern web designs, we integrate simulated human-like interactions to navigate pages. Collection quality is further improved through filtering strategies combining rule-based heuristics and LLM-based evaluation, effectively pruning low-quality links while increasing information density. Additional safety constraints are enforced to prevent the collection of sensitive or restricted information, as detailed further in Appendix A.

\subsection{KG Construction}

We construct a KG to transform unstructured and semi-structured content into a computable, reasoning-friendly semantic space. Formally, we define the KG as $G = (V, E, R)$, where \(V\) is the set of nodes, \(E\) is the set of edges, and \(R\) is the set of relation types.

\vspace{-0.3cm}
\paragraph{Node Extraction.} 
Nodes are extracted by parsing a document or webpage \(D\) to identify elements such as paragraphs, headings, hyperlinks, forms, buttons, and table cells. Each element is mapped to a node:  
\[
V = \{ v_i \mid v_i \in \text{Elements}(D), \; \text{type}(v_i) \in \textit{NodeTypeSet} \},
\]  
where \(\textit{NodeTypeSet}\) includes Paragraph, Heading, Hyperlink, Form, Button, Table, and other domain-specific elements. The contextual path \(\text{Path}(v_i)\) is preserved to maintain the DoM hierarchy or document structure.

\vspace{-0.3cm}
\paragraph{Node Representation.} 
Each node \(v_i\) contains textual content \(c_i^T\) (\textit{e.g.}, text, captions, alt text) and visual content \(c_i^V\) (\textit{e.g.}, images, screenshots). Visual content is first converted to textual descriptions using a function \(\phi_{\text{visual}}\):  
\[
c_i^{T+V} = c_i^T \,||\, \phi_{\text{visual}}(c_i^V),
\]  
where \(||\) denotes text concatenation. The combined textual representation is then embedded into a vector space:  
\[
h_i = f_{\text{embed}}(c_i^{T+V}), \quad f_{\text{embed}}: \text{Text} \rightarrow \mathbb{R}^d,
\]  
where \(d\) is the embedding dimension (\textit{e.g.}, \(d=384\) for all-MiniLM-L6-v2). The resulting vector \(h_i\) is stored in a vector database \(D_{\text{vec}}\) for efficient semantic search and similarity matching.

\vspace{-0.3cm}
\paragraph{Edge Construction.} 
Relations between nodes are captured as a heterogeneous edge set:  
\[
E = E_{\text{text}} \cup E_{\text{web}}, \quad
E \subseteq V \times V \times R,
\]  
where \(E_{\text{text}}\) encodes text-based relationships, including structural relations (\textit{e.g.}, sequence, contains), semantic associations (\textit{e.g.}, entity relations, semantic similarity), contextual relations (\textit{e.g.}, figure or table context), and reference relations (\textit{e.g.}, co-reference, cross-document links). \(E_{\text{web}}\) models web-specific interactions, such as navigation relations (\textit{e.g.}, page navigation, form submission), interaction relations (\textit{e.g.}, click triggers), and layout relations (\textit{e.g.}, visual layout or data flow).

\subsection{Subgraph Sampling}

In the subgraph sampling stage, given a task objective $g$, \textsc{Graph2Eval} extracts a local subgraph $G_g=(V_g,E_g,R)\subseteq G$ by selecting relevant nodes and their interconnections.
\vspace{-0.3cm}
\paragraph{Scenario-Specific Sampling.} 
The subgraph sampling follows different strategies depending on the scenario: \textbf{(1)~Document Comprehension:} Nodes include \textit{DocumentElementNode} (paragraph, heading, table, image), \textit{EntityNode} (person, location, organization), and \textit{SemanticChunkNode}, capturing both semantic content and structural roles. Sampling prioritizes semantic relevance (via embeddings) and structural coherence (via $StructMatch$). Only nodes of the relevant types are included. \textbf{(2)~Web Interaction:} Sampling follows a seed-driven strategy. First, task-specific seed nodes $S_{\text{seed}}(g)$ (buttons, forms, navigation links) are identified. Second, the $k$-hop neighbors of each seed node are collected, including valid \textit{WebPageNode} and \textit{WebElementNode} entities. This ensures that the local interaction context is captured around the seeds.

\vspace{-0.3cm}
\paragraph{Sampling Workflow.} 
We present the workflow of subgraph sampling in Algorithm~\ref{alg:sampling}. For notation, the KG is denoted as $G=(V,E,R)$, and the extracted subgraph is $G_g=(V_g,E_g,R)$, where $V_g \subseteq V$ and $E_g \subseteq E$. The task objective is $g$, which is mapped to an embedding $h_g$ using the same embedding function $f_{\text{embed}}(\cdot)$ used for nodes. Each node $v_i$ is similarly represented by its embedding $h_i$. In the document mode, $\tau$ denotes the cosine similarity threshold for node relevance ($\cos(h_i, h_g)$), and $StructMatch(\cdot)$ is the function for checking structural alignment. In the web mode, $S_{\text{seed}}(g)$ represents the set of task-specific seed nodes, and $k$ is the neighborhood hop distance for collecting context (via $\textsc{Neighbor}(\cdot)$). Node sets are further constrained by $NodeTypeSet$ (document) or $WebNodeSet$ (web) filters.

\begin{algorithm}[t]
\small
\DontPrintSemicolon
\SetAlgoLined
\LinesNumbered
\caption{Workflow of Subgraph Sampling}
\label{alg:sampling}
\KwIn{
    KG $G=(V,E,R)$, task objective $g$, mode $m \in \{\text{document}, \text{web}\}$, \\
    threshold $\tau$ (document), neighborhood $k$ (web), seed nodes $S_{\text{seed}}(g)$ (web)
}
\KwOut{Sampled subgraph $G_g=(V_g,E_g,R)$}

$V_g \gets \emptyset$, $E_g \gets \emptyset$ 

\ForEach{$v_i \in V$}{
    \tcc{Evaluate node $v_i$ based on current mode (document or web)}

    \If{$m = \text{document}$}{
        $h_i \gets \textsc{Embedding}(v_i)$ 

        \If{$\cos(h_i, h_g) > \tau$ \textbf{or} StructMatch$(v_i,g)=1$}{
            \If{$v_i \in NodeTypeSet$}{
                $V_g \gets V_g \cup \{v_i\}$ 
            }
        }
    }

    \If{$m = \text{web}$}{
        \If{$v_i \in S_{\text{seed}}(g)$}{
            $V_g \gets V_g \cup \{v_i\}$ 

            $N_i \gets \textsc{Neighbor}(v_i,k)$ 

            \ForEach{$v_j \in N_i$}{
                \If{$v_j \in WebNodeSet$}{
                    $V_g \gets V_g \cup \{v_j\}$ 
                }
            }
        }
    }
}
$E_g \gets \{ (v_i,v_j) \in E \mid v_i, v_j \in V_g \}$ 

\tcc{Final subgraph $G_g$ contains selected nodes and their interconnections}
\Return{$G_g = (V_g, E_g, R)$}
\end{algorithm}

\subsection{Task Generation} 
In \textsc{Graph2Eval}, we use subgraphs from KGs to generate tasks. Each subgraph is transformed into an executable and evaluable task, 
yielding two task types: document understanding and web interaction.

\paragraph{Document Understanding.} For document understanding tasks, the generation pipeline of \textsc{Graph2Eval} consists of four stages: \textbf{(1)~Task Templates}: We maintain a library of \emph{task templates} that cover fundamental categories such as question answering, comparison, analysis, and reasoning (see Appendix B for details). \textbf{(2)~Subgraph Sampling}: \textsc{Graph2Eval} performs \emph{subgraph sampling} from the KG to select subgraphs that satisfy the constraints imposed by the task template. The sampled subgraph must meet template-specified requirements, including mandatory node and edge types, node count ranges, and maximum hop distance. By adjusting subgraph size, edge types, and sampling strategies, the framework can flexibly control task complexity and reasoning depth. \textbf{(3)~Variable Extraction}: From the sampled subgraph, \textsc{Graph2Eval} extracts template variables such as node contents, edge relations, contextual information, and metadata, which serve as necessary inputs for task generation. \textbf{(4)~Task Generation}: Given the subgraph, \textsc{Graph2Eval} combines the structural content with the template-defined context, and further leverages LLMs to generate concrete task instances.

\paragraph{Web Interaction.} For web interaction tasks, we propose a \emph{Seed-Driven Subgraph Sampling Strategy}, which consists of four stages: \textbf{(1) Task Seed Identification}: \textsc{Graph2Eval} first parses the page to identify key operational nodes (\textit{e.g.}, buttons, input boxes, forms, navigation links) as ``task seeds,'' thereby anchoring task execution to actual page functionality. \textbf{(2) Subgraph Sampling}: Based on these seeds, relevant contextual nodes and interaction edges are sampled from the KG to construct a subgraph. \textbf{(3) Meta-path Matching}: Meta-path patterns are then applied to match and extend the subgraph, producing concrete task chains. Details of the meta-path design and implementation are provided in the Appendix C.  
\textbf{(4) Dynamic Task Generation}: Once the subgraph and task chain are obtained, LLMs generate concrete task instances by combining the subgraph structure, meta-paths, and page context information (\textit{e.g.}, screenshots and element lists processed by the Set of Marks). For example, if only a search box, submit button, and result items are detected, the task chain becomes \textit{Search + Detail}; if a filter is also present, the chain becomes \textit{Search + Filter + Detail}.  
The \emph{seed $\rightarrow$ subgraph sampling $\rightarrow$ meta-path matching} pipeline drives task generation with contextual relevance, enables multi-hop and branching flexibility through diverse sampling strategies, and offers controllability via seed selection and meta-path design. This compositional mechanism avoids rigid \textit{all-or-nothing} constraints.

% \begin{figure}[h]
%     \centering
%     \includegraphics[width=0.45\textwidth]{figures/task_coverage_chart.pdf}
%     \caption{Coverage proportions of Web and Doc task dimensions used in the task optimization.}
%     \label{fig:coverage_optimization}
% \end{figure}

% For safety-oriented task generation, we adopt a \textit{policy-aware} approach tailored to textual data. External policy files are parsed to extract safety constraints (\textit{e.g.}, privacy protection, access control, malicious input defense), and LLM-based reverse parsing and semantic perturbation are employed to generate adversarial inputs embedded into the textual corpus. Example tasks include “identify forged bibliographic entries” or “detect tampered numerical values,” evaluating agent robustness under adversarial scenarios. Web safety tasks, however, are inherently different due to dynamic threats in interactive environments (\textit{e.g.}, phishing buttons, fake forms), and thus are not generated offline. In \textsc{Graph2Eval}, web security evaluation occurs naturally through real-world interaction, maintaining fidelity to actual threat models.

\begin{tcolorbox}[custombox={green!30!black}{Highlights}]
The divergence between doc tasks and web tasks stems from their execution paradigms: doc tasks require an agent to perform a limited number of API calls within a few dialogue turns, whereas web tasks involve sequential, multi-step interactions within dynamic web environments, thereby necessitating distinct generation strategies.
\end{tcolorbox}

\subsection{Coverage Optimization}

We use a multi-stage optimization framework to ensure the quality, diversity, and representativeness of generated tasks. 
For web interaction tasks, candidate tasks are first filtered using either LLM-based or rule-based quality scores, and coverage is quantified across node type, edge type, pattern, page-level, website type, and difficulty dimensions. Novelty is measured via multi-level similarity, and tasks are iteratively selected using a Maximal Marginal Relevance (MMR)–based strategy. For document understanding tasks, coverage emphasizes semantic diversity across task type, difficulty, template, variable, and content length, with novelty assessed via LLM-based semantic similarity and selection also guided by MMR. 

% The final task set balances quality, coverage, and diversity, as shown in Figure~\ref{fig:coverage_optimization}.

\section{Experiments}

This section presents experiments on \textsc{Graph2Eval} and \textsc{Graph2Eval-Bench}, evaluating task semantic consistency, solvability, and discriminative ability. Limitations of \textsc{Graph2Eval} are discussed in Appendix D.

\label{sec:experiment}

\subsection{Implementation Details}
\paragraph{Agents.}
We evaluate \textsc{Graph2Eval} on two types of agents: RAG Agents and Web Agents. The RAG Agents include both Single Agent and Multi-Agent setups for document understanding, with the Multi-Agent architecture comprising a planner, retriever, reasoner, verifier, and summarizer. Web Agents are responsible for multi-step web interactions, including the Set-of-Marks (SoM) Agent~\citep{yang2023setofmarkpromptingunleashesextraordinary}, which leverages \textit{SoM-annotated images} to provide richer multimodal context, and Agent S 2.5~\citep{agashe2025agents2compositionalgeneralistspecialist}, which incorporates \textit{task-aligned reflection} and \textit{memory management} to enhance reasoning. Detailed agent architectures are provided in Appendix E.

\paragraph{Models \& Baselines.}
Task generation and optimization are based on \texttt{GPT-4o}~\citep{hurst2024gpt} and we evaluate multiple model families: GPT Series~\citep{hurst2024gpt} (\texttt{GPT-4o}, \texttt{GPT-4o mini}, \texttt{GPT-4.1 mini}), Deepseek Series~\citep{guo2025deepseek} (\texttt{Deepseek-V3}, \texttt{Deepseek-V3.1}), Qwen Series~\citep{bai2023qwen} (\texttt{Qwen2.5-VL-7B}, \texttt{Qwen2.5-VL-32B}, \texttt{Qwen2.5-VL-72B}), and Gemini Series~\citep{deepmind2025gemini25flash} (\texttt{Gemini 2.5 Flash}), as well as the GUI understanding model \texttt{UI-TARS-1.5-7B}~\citep{qin2025uitarspioneeringautomatedgui}. All models use a temperature of 0.1, with \texttt{all-MiniLM-L6-v2} embeddings for graph construction. In addition, we compare \textsc{Graph2Eval} with the KG-free variant, as well as with OSWorld~\cite{OSWorld}, MMBench~\cite{liu2024mmbenchmultimodalmodelallaround}, and TaskCraft~\cite{shi2025taskcraftautomatedgenerationagentic}, to evaluate differences across task generation strategies.

\paragraph{Metrics.} 
We evaluate document understanding tasks using three metrics: (1)~\textit{F1}, (2)~\textit{ROUGE-L}, and (3)~\textit{LLM-as-a-Judge} (abbreviated as \textit{LLM Judge}), capturing performance at both the rule-based and semantic understanding levels. For web tasks, success is measured by the Success Rate (SR), defined as the proportion of tasks judged successfully completed by an LLM evaluator. The reliability of this metric was validated against human annotations (Appendix F.3). Additionally, we define metrics for task quality as follows: (1)~\textit{Consistency}, measuring the proportion of tasks whose content is semantically aligned with the source data, and (2)~\textit{Solvability}, measuring the proportion of tasks that can be successfully completed given the provided context. Detailed formulas are provided in the Appendix F.

\begin{figure}[h]
    \centering
    \includegraphics[width=0.48\textwidth]{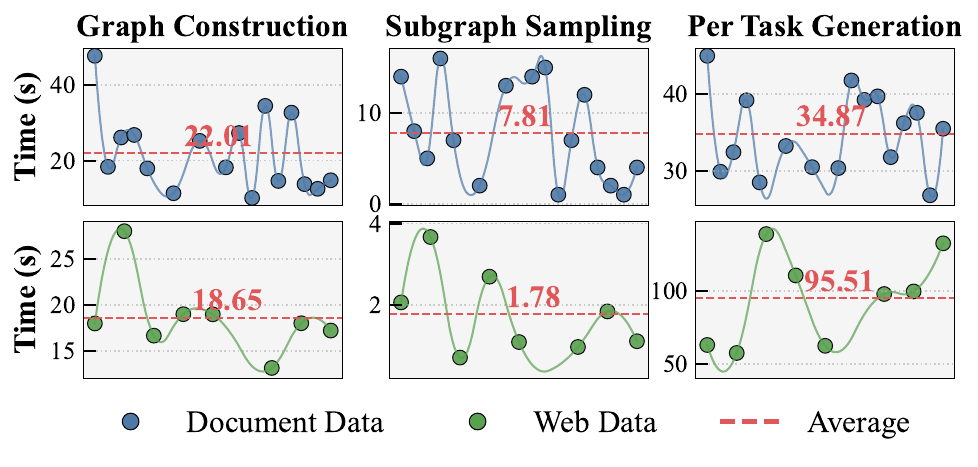}
    \caption{Comparison of \textsc{Graph2Eval} Processing Time on Document and Web Data.}
    \label{fig:dataset_stage_time}
\end{figure}

% % ==== 图1 ====
% \begin{figure}[t]
%     \centering
%     \includegraphics[width=\linewidth]{figures/heatmap.pdf}
%     \caption{The heatmap shows the LLM evaluation quality scores of model responses across different task types (X-axis) and models (Y-axis).}
%     \label{fig:heatmap}
% \end{figure}

% % ==== 图2 ====
% \begin{figure}[t]
%     \centering
%     \includegraphics[width=\linewidth]{figures/grouped_bar_two_metrics_gradient_models.pdf}
%     \caption{The grouped bar chart shows \textit{F1} and \textit{ROUGE-L} scores across different task types (X-axis) and models. Each model uses gradient-filled bars to reflect the score magnitude.}
%     \label{fig:bar_chart}
% \end{figure}

\subsection{Main Results}

\paragraph{Dataset Construction.}
The details about \textsc{Graph2Eval-Bench} are presented in Table~\ref{tab:dataset_construction}. For document understanding tasks, high-quality papers were collected as the data source, while Web interaction tasks drew data from various websites, including screenshots and DOM structures. To validate the constructed graph, we randomly sampled 100 node pairs from both data sources and manually verified the correctness of nodes and edges, with the high KG Edge Precision confirming the graph’s reliability. Furthermore, Figure~\ref{fig:dataset_stage_time} shows that \textsc{Graph2Eval} requires substantially less time for task generation compared to manual construction. A detailed dataset composition is provided in Appendix G.

\begin{table}[h]
    \centering
    \caption{Statistics of \textsc{Graph2Eval-Bench}.}
    \begin{tabular}{l|c}
        \toprule
        \textbf{Metric} & \textbf{Value (Average)} \\
        \midrule
        Documents / Websites & 16 / 8 \\
        Tasks (Doc / Web) & 1002 / 317 \\
        Tasks per document / website & 83.5 / 48.4 \\
        Task types (Doc / Web) & 12 / 7 \\
         KG Edge Precision & 88\% \\
        \bottomrule
    \end{tabular}
    \label{tab:dataset_construction}
\end{table}

\begin{table*}[t]
\centering
\setlength{\tabcolsep}{2.5pt}
\renewcommand{\arraystretch}{1.1}
\small
\caption{Performance comparison of models under single agent and multi-agent evaluation settings. Best and second-best values are \textbf{bolded} and \underline{underlined}, respectively.}
\begin{tabular}{l c c c c c c c c}
\toprule
\multirow{2}{*}{\textbf{Model}} 
& \multicolumn{4}{c}{\textbf{Single Agent}} 
& \multicolumn{4}{c}{\textbf{Multi-Agent}} \\
\cmidrule(lr){2-5} \cmidrule(lr){6-9}
& \textit{F1} & \textit{Rouge-L} & \textit{LLM Judge} & Avg. Token
& \textit{F1} & \textit{Rouge-L} & \textit{LLM Judge} & Avg. Token \\
\midrule
\rowcolor{CadetBlue!20} \multicolumn{9}{c}{\textbf{Multimodal Model}} \\
\addlinespace[0.7mm]
 \textbf{\texttt{GPT-4o}}  
& \textbf{0.5766} & \textbf{0.4874} & 0.7854 & \textbf{1631.82}
& \textbf{0.5916} & \textbf{0.4873} & 0.7623 & 3560.92 \\
\textbf{\texttt{GPT-4.1 mini}} 
& 0.4202 & 0.4202 & \underline{0.8202} & 3593.13
& 0.3496 & 0.3068 & 0.6972 & 4679.13\\
\textbf{\texttt{Qwen2.5-VL-72B}}  
& \underline{0.5730} & \underline{0.4837} & 0.7094 & 2394.19
& \underline{0.5673} & \underline{0.4711} & 0.6999 & 3855.65 \\
\textbf{\texttt{Qwen2.5-VL-32B}}  
& 0.3677 & 0.3300 & 0.4811 & 2275.01
& 0.4341 & 0.3813 & 0.5008 & 3779.32 \\ 
\textbf{\texttt{Qwen2.5-VL-7B}}   
& 0.2093 & 0.1939 & 0.5427 & 2455.17
& 0.3496 & 0.2548 & 0.6973 & 3732.27 \\
\midrule
\rowcolor{CadetBlue!20} \multicolumn{9}{c}{\textbf{Textual Model}} \\
\addlinespace[0.7mm]
\textbf{\texttt{Deepseek-V3}} 
& 0.5376 & 0.4518 & \textbf{0.8351} & \underline{1710.65}
& 0.5497 & 0.4635 & \textbf{0.7984} & \underline{3462.88} \\
\textbf{\texttt{Deepseek-V3.1}}  
& 0.5276 & 0.4329 & 0.7816 & 1777.98
& 0.5141 & 0.4253 & \underline{0.7875} & \textbf{3435.12} \\
\bottomrule
\end{tabular}%
\label{tab:single_multi_agent_perf}
\end{table*}

% \begin{table}[h]
%   \centering
%   \small
%   \setlength{\tabcolsep}{1.5pt}
%   \caption{Benchmark Comparison.}
%   \label{tab:datasets-binary}
%   \begin{tabular}{lcccc}
%     \toprule
%     \textbf{Aspect} & \textbf{OSWorld~\cite{OSWorld}} & \textbf{GAIA~\cite{mialon2023gaiabenchmarkgeneralai}} & \textbf{TaskCraft~\cite{shi2025taskcraftautomatedgenerationagentic}} & \textbf{Our} \\
%     \midrule
%     Task Type & OS & text & text & OS,text \\
%     label   & human & human & auto & auto \\
%     % Task Num. & 369  & 466 & 20,000 & 1,319 \\
%     Env. Type & Simulation  & N/A & N/A & Real Web \\
%     Interaction &  & chat & chat &  \\
%     \bottomrule
%   \end{tabular}
% \end{table}

\begin{table*}[t]
\centering
\small
\renewcommand{\arraystretch}{1.1}
\caption{Performance comparison of various models under \textit{SoM Agent} and \textit{Agent S 2.5} evaluation settings. Performance is measured as \textit{SR} of tasks. Best and second-best values are \textbf{bolded} and \underline{underlined}, respectively.}
\begin{tabular}{lccccccc|>{\columncolor{yellow!30}}c}
\toprule
\textbf{Model} & \textit{Basic Nav.} & \textit{Toast} & \textit{Cont.} & \textit{Search} 
& \textit{Modal} & \textit{Buss. Navi.} & \textit{Button} & \textbf{Overall} \\
\midrule
\rowcolor{CadetBlue!20} \multicolumn{9}{c}{\textbf{SoM Agent}} \\
\addlinespace[0.7mm]
\textbf{\texttt{Gemini 2.5 Flash}} & 0.1778 & 0.0000 & 0.2000 & 0.1134 & 0.0000 & 0.2500 & 0.1875 & 0.1451 \\
\textbf{\texttt{GPT-4o mini}} & 0.0667 & 0.0000 & 0.0000 & 0.1237 & 0.0000 & 0.0750 & 0.0000 & 0.0946 \\
\textbf{\texttt{Qwen2.5-VL-72B}} & 0.1333 & \textbf{1.0000} & 0.2000 & 0.1031 & 0.1666 & \underline{0.2750} & 0.1250 & 0.1388 \\
 \textbf{\texttt{Qwen2.5-VL-32B}} & 0.0222 & \textbf{1.0000} & 0.0667 & 0.0155 & 0.0000 & 0.0750 & 0.0000 & 0.0283 \\
 \textbf{\texttt{Qwen2.5-VL-7B}} & 0.0000 & 0.0000 & 0.0000 & 0.0103 & 0.0000 & 0.0000 & 0.0000 & 0.0063 \\
 \textbf{\texttt{UI-TARS-1.5-7B}} & 0.0667 & 0.0000 & 0.0000 & 0.0773 & 0.0000 & 0.0750 & 0.1250 & 0.0491 \\

\midrule
\rowcolor{CadetBlue!20} \multicolumn{9}{c}{\textbf{Agent S 2.5}} \\
\addlinespace[0.7mm]
\textbf{\texttt{Gemini 2.5 Flash}} & \textbf{0.4889} & \textbf{1.0000} & \textbf{0.6667} & \textbf{0.6340} & \textbf{1.0000} & \textbf{0.5500} & \textbf{0.6250} & \textbf{0.6920} \\
 \textbf{\texttt{GPT-4o mini}} & \underline{0.3334} & 0.0000 & 0.2000 & 0.3763 & 0.5000 & 0.1750 & 0.2500 & 0.3312 \\
 \textbf{\texttt{Qwen2.5-VL-72B}} & \underline{0.3334} & \textbf{1.0000} & \underline{0.2667} & \underline{0.4278} & \underline{0.8333} & 0.1500 & \underline{0.5625} & \underline{0.3880} \\
  \textbf{\texttt{Qwen2.5-VL-32B}} & 0.1778 & 0.0000 & 0.2000 & 0.1546 & 0.3333 & 0.0750 & 0.1250 & 0.1514 \\
\textbf{\texttt{Qwen2.5-VL-7B}} & 0.1112 & 0.0000 & 0.0666 & 0.1392 & 0.1667 & 0.0750 & 0.0000 & 0.1167 \\
 \textbf{\texttt{UI-TARS-1.5-7B}} & 0.0889 & 0.0000 & 0.0667 & 0.0979 & 0.0000 & 0.1250 & 0.1250 & 0.0719 \\

\bottomrule
\end{tabular}
\label{tab:web_analysis}
\end{table*}

\paragraph{Analysis of Document Understanding Tasks.}
 We evaluated \textsc{Graph2Eval-Bench} under both single agent and multi-agent collaboration settings, with results in Table \ref{tab:single_multi_agent_perf}. We tested both multimodal LLMs and textual LLMs (in text mode, \textsc{Graph2Eval} converts figure interpretation tasks into text-based reasoning tasks by extracting image metadata, including titles, captions, alt text, and OCR-extracted text). Overall, \texttt{GPT-4o} achieves the highest \textit{F1} and \textit{ROUGE-L} scores, while its performance under \textit{LLM-as-a-Judge} ranks second. On the other hand, \texttt{Deepseek-V3} performs best in the  \textit{LLM-as-a-Judge}. 
 
 % Notably, multi-agent collaboration does not lead to significant improvements, suggesting that this multi-agent design provides limited enhancement for RAG-based document understanding. 

% \paragraph{Task Type Performance.} Figures \ref{fig:heatmap} and \ref{fig:bar_chart} present single agent evaluation results across different task categories. \texttt{Deepseek-V3} and \texttt{GPT-4o} consistently achieve top or near-top performance across all types, and clear differences are observed among models of varying parameter scales. These findings indicate that the tasks generated by \textsc{Graph2Eval} are sufficiently discriminative and challenging, effectively capturing differences in language understanding and reasoning capabilities across models.

% \begin{tcolorbox}[custombox={blue!55!black}{Findings}]
% Our generated datasets highlight efficiency bottlenecks in multi-agent collaboration on document understanding tasks. Higher token costs do not improve overall performance and can slightly degrade results.
% \end{tcolorbox}

\paragraph{Analysis of Web Interaction Tasks.}
We evaluated the web interaction tasks in the \textsc{Graph2Eval-Bench} using SoM Agent and Agent S 2.5, with the results presented in Table \ref{tab:web_analysis}. Overall, Agent S 2.5 consistently outperforms SoM Agent across nearly all task types. On \texttt{Gemini 2.5 Flash}, both agents achieved their best overall performance, with SoM Agent reaching 14.51\% and Agent S 2.5 achieving 69.20\%, demonstrating a clear performance gap. The open-source model \texttt{Qwen2.5-VL-72B} also achieved strong results, ranking as the second-best overall. By contrast, \texttt{GPT-4o mini} showed competitive performance on specific tasks (\textit{e.g.}, business navigation and modal interaction), but its overall effectiveness remained limited. The relatively low scores of \texttt{Qwen2.5-VL-32B}, \texttt{Qwen2.5-VL-7B}, and \texttt{UI-TARS-1.5-7B} demonstrate that the benchmark effectively distinguishes performance differences across models of varying scales. A detailed case study is provided in Appendix H.
\begin{tcolorbox}[custombox={blue!55!black}{Findings}]
The performance gap between Agent S 2.5 and SoM Agent shows that our tasks emphasize multi-step reasoning rather than simple visual grounding, effectively assessing the advanced reasoning, reflection, and memory capabilities present in Agent S 2.5 but absent in SoM Agent.
\end{tcolorbox}

% \subsection{Comparison with Existing Datasets}

% \begin{table}[t]
% \centering
% \small
% \renewcommand{\arraystretch}{1.2}
% % \setlength{\tabcolsep}{3.6pt}
% \caption{Performance comparison of various models under \textit{SoM Agent} and \textit{Agent S 2.5} evaluation settings. Performance is measured as \textit{SR} of tasks. Best and second-best values are \textbf{bolded} and \underline{underlined}, respectively.}
% \begin{tabular}{lcc}
% \toprule
% \textbf{Benchmark} & \textit{Acc} & \textit{Toast} \\
% \midrule
% \rowcolor{CadetBlue!20} \multicolumn{3}{c}{\underline{\textbf{Document Understanding}}} \\
% \addlinespace[0.7mm]
% \textbf{\texttt{GAIA}} & 0.1778 & 0.0000  \\
%  \textbf{\texttt{Graph2Eval-Bench}} & 0.0667 & 0.0000  \\
% \midrule
% \rowcolor{CadetBlue!20} \multicolumn{3}{c}{\underline{\textbf{Web Interaction}}} \\
% \addlinespace[0.7mm]
% \textbf{\texttt{gemini-2.5-flash}} & \textbf{0.4889} & \textbf{1.0000} \\
%  \textbf{\texttt{gpt-4o-mini}} & \underline{0.3334} & 0.0000 \\

% \bottomrule
% \end{tabular}
% \label{tab:dataset_comparison}
% \end{table}

% \begin{figure}[t]
%     \centering
%     \includegraphics[width=0.45\textwidth]{figures/case_study4.pdf}
%     \caption{Case Study.}
%     \label{fig:method}
% \end{figure}

\subsection{Ablation Studies}
To evaluate the role of the KG as the task space on the quality of generated tasks, we compare \textsc{Graph2Eval} with a KG-free variant that generates tasks directly from processed documents and web data. Both methods produce 1000 tasks using the same data set with \texttt{GPT-4o} (500 document understanding tasks and 500 web interaction tasks).

For human evaluation, the generated tasks were independently reviewed by two human annotators, who assessed each task based on two criteria: \textit{Consistency} and \textit{Solvability}. Final scores were computed as the proportion of cases where both annotators reached agreement. 
In addition to human evaluation, we also conduct agent evaluation using \texttt{Gemini 2.5 Flash} under the same agent configuration. The results are shown in Table~\ref{tab:ablation_studies}. From both human and agent evaluations, we observe: 
(1)~Tasks generated by \textsc{Graph2Eval} reveal clearer performance gaps between models of different sizes. 
(2)~The KG-free variant produces web tasks that are largely confined to single-page interactions, and multi-page workflows are often unsolvable due to missing inter-page relations. 
(3)~Tasks generated without a KG require LLM or manual review to resolve ambiguities, whereas \textsc{Graph2Eval} leverages KGs to enforce semantic consistency and ensure task solvability.

\begin{table}[ht]
  \centering
  \small
  \caption{Ablation studies on \textsc{Graph2Eval}.}
  \renewcommand\arraystretch{1.1} 
  \label{tab:method-comparison}
  \resizebox{\linewidth}{!}{%
  \begin{tabular}{l|cc|cc}
    \toprule
    \textbf{Model}
    & \multicolumn{2}{c|}{\textbf{\textsc{Graph2Eval} w/o KG}}
    & \multicolumn{2}{c}{\textbf{\textsc{Graph2Eval}}} \\
    \cmidrule(lr){2-3} \cmidrule(lr){4-5}
    & Doc & Web & Doc & Web \\
    \midrule
    \rowcolor{CadetBlue!20}\multicolumn{5}{c}{\textbf{Agent Evaluation}}\\
    \textbf{\texttt{Qwen2.5-VL-72B}} & 0.68 & 0.12 & 0.85 & 0.24 \\
    \textbf{\texttt{Qwen2.5-VL-32B}} & 0.36 & 0.08 & 0.50 & 0.17 \\
    \textbf{\texttt{Qwen2.5-VL-7B}} & 0.29 & 0.07 & 0.44 & 0.12 \\
    \midrule
    \rowcolor{CadetBlue!20}\multicolumn{5}{c}{\textbf{Human Evaluation}}\\
    \textit{Consistency} & 0.74 & 0.62 & 0.95 & 0.78 \\
    \textit{Solvability} & 0.73 & 0.60 & 0.93 & 0.72
    \\
    \bottomrule
  \end{tabular}
  }
\label{tab:ablation_studies}
\end{table}

\subsection{Comparison with Existing Methods}
\paragraph{Human-Annotated Data Construction Methods.}
We compare the characteristics of \textsc{Graph2Eval} with two representative benchmarks: OSWorld, which focuses on web interaction capabilities, and the multimodal visual question answering benchmark MMBench. Compared with existing human-annotated datasets, \textsc{Graph2Eval} enables automated construction of tasks with low time cost, while maintaining strong semantic consistency and solvability, and covering a broader range of task types.

\begin{table}[h]
\centering
\caption{Comparison on OSWorld, MMBench and \textsc{Graph2Eval}. N/A denotes Not Applicable.}
\renewcommand\arraystretch{1.2} 
\resizebox{1\linewidth}{!}{%
\begin{tabular}{l|c|c|c}
\toprule
\textbf{Metric} & \textbf{OSWorld}~\cite{OSWorld} & \textbf{MMBench}~\cite{liu2024mmbenchmultimodalmodelallaround} & \textbf{\textsc{Graph2Eval}} \\
\midrule
Data Source & Human + scripted & Curated QA & Synthetic \\
Task Types & Real web/desktop & Multimodal QA & Doc + Web \\
Time Cost & High & Moderate & Low \\
Consistency & High & High & High \\
Solvability & High & N/A & High \\
\bottomrule
\end{tabular}%
}
\label{tab:dataset_compare}
\end{table}

\vspace{-0.3cm}
\paragraph{Synthetic Data Construction Methods.} 
We compare TaskCraft and \textsc{Graph2Eval} in Table~\ref{tab:taskcraft_compare}. 
TaskCraft uses a bottom-up approach, first generating atomic tasks and then combining them into complex ones from raw documents, 
while \textsc{Graph2Eval} follows a top-down paradigm, constructing a complete knowledge graph and sampling subgraphs to form tasks based on intrinsic relationships within the data. Moreover, while TaskCraft focuses on multi-tool use, \textsc{Graph2Eval} generates a broader range of task types.

% \begin{table}[ht]
% \centering
% \small
% \setlength{\tabcolsep}{1.5pt}
% \caption{Performance Comparison on OSWorld, MMBench and \textsc{Graph2Eval-Bench}. N/A denotes Not Applicable.}
% \resizebox{\linewidth}{!}{%
% \begin{tabular}{lccc}
% \toprule
% \textbf{Model} & \textbf{OSWorld} & \textbf{MMBench} & \textbf{\textsc{Graph2Eval-Bench}} \\
% \midrule
% \rowcolor{CadetBlue!20}\multicolumn{4}{c}{\textbf{Performance on Web Datasets}} \\
% \textbf{\texttt{Qwen2.5-VL-72B}}      & 0.050 & N/A     & 0.133  \\
% \textbf{\texttt{qwen2.5-VL-32B}}      & 0.039 & N/A    & 0.022  \\
% \textbf{\texttt{UI-TARS-1.5-7B}}      & 0.273  & N/A   & 0.097     \\
% \midrule
% \rowcolor{CadetBlue!20}\multicolumn{4}{c}{\textbf{Performance on Doc Datasets}}\\
% \textbf{\texttt{GPT-4o}}              & N/A     & 0.860  & 0.785 \\
% \textbf{\texttt{Qwen2.5-VL-72B}}      & N/A     & 0.882 & 0.709 \\
% \textbf{\texttt{Qwen2.5-VL-32B}}      & N/A     & 0.840 & 0.481 \\
% \bottomrule
% \end{tabular}%
% }
% \label{tab:dataset_compare}
% \end{table}

% \vspace{-0.5cm}

\begin{table}[h!]
\centering
\renewcommand\arraystretch{1.3} 
\caption{Comparison between \textsc{Graph2Eval} and TaskCraft.}
\resizebox{\linewidth}{!}{%
\begin{tabular}{l|l|l} 
\toprule
\textbf{Metric} & \textbf{TaskCraft}~\cite{shi2025taskcraftautomatedgenerationagentic} & \textbf{\textsc{Graph2Eval}} \\
\midrule
Methodology & Bottom-up generation & Top-down generation \\
\cmidrule(lr){1-1} \cmidrule(lr){2-3}
Task Space & Processed documents & Knowledge graphs \\
\cmidrule(lr){1-1} \cmidrule(lr){2-3}
Task Complexity & \makecell[l]{Composition of \\atomic tasks} & \makecell[l]{Exploitation of intrinsic\\data relationships} \\
\cmidrule(lr){1-1} \cmidrule(lr){2-3}
Task Types & Multi-tool use & \makecell[l]{Doc Understanding, Web} \\
\bottomrule
\end{tabular}%
}
\label{tab:taskcraft_compare}
\end{table}

% \subsection{Discussion and Limitation}
% The "cheap but scalable" feature of \textsc{Graph2Eval} enables rapid agent evaluation in data-scarce domains. Our experiments show that even advanced models fail on tasks with simple node-combination variants, revealing that current models still lack robustness to minor task perturbations, a weakness often masked by static, manually designed evaluations. Consequently, static benchmarks may overestimate model capabilities, while dynamic and scalable tasks provide a more faithful measure of generalization. This leads to a key insight for benchmark design: an ideal multimodal or structured understanding benchmark should combine scalability, controllable perturbations, and challenge in order to expose the latent weaknesses of models.
\section{Conclusion and Future Work}

In this work, we introduce \textsc{Graph2Eval}, an automatic task generation framework that leverages KGs as an intermediate representation. By systematically modeling entities and their relationships within documents and web data, the framework integrates multi-source data into a unified task space, enabling scalable, semantically consistent, and solvable agent task creation for evaluating agent capabilities across diverse scenarios. Based on this framework, we constructed the \textsc{Graph2Eval-Bench} dataset. Experimental results demonstrate that tasks generated by \textsc{Graph2Eval} exhibit improved semantic consistency and solvability, effectively cover a wide range of scenarios, and reliably assess document understanding and web interaction capabilities across different settings.

Future research will explore the following directions: (1) incorporating formalized safety policies to generate testable tasks for evaluating agent robustness under adversarial prompts, malicious environments, and other challenging scenarios; and (2) exploiting the structural properties of KGs to localize errors at the level of nodes and edges, providing fine-grained, interpretable insights into agents’ weaknesses in multimodal document understanding and web interaction task performance.

\paragraph{Acknowledgements.}
This work is supported by the National Natural Science Foundation of China (No. 62402429, U24A20326, 62441236), the Key Research and Development Program of Zhejiang Province (No. 2025C01026), the Ningbo Yongjiang Talent Introduction Programme (2023A-397-G), and the Young Elite Scientists Sponsorship Program by CAST (2024QNRC001). This work was supported by MYbank, Ant Group. We also gratefully acknowledge the support of the Zhejiang University Education Foundation Qizhen Scholar Foundation. 
{
    \small
    \bibliographystyle{ieeenat_fullname}
    \bibliography{main}
}

\clearpage
\appendix
\section{Task Content Constraints}
\label{sec:task_content_constraints}
For document understanding tasks, the content safety is relatively controllable due to the constraints of task type and information sources. In contrast, web interaction tasks require the system to autonomously collect information from web pages, so we have implemented a multi-stage filtering mechanism to ensure safety.

\paragraph{Data Parsing Stage.}
We utilize LLMs to evaluate both web links and page content. During this process, the LLM evaluator is explicitly instructed to avoid pages containing personal or private information (e.g., pages with email addresses, phone numbers, or physical addresses) and to prioritize high-quality pages that do not involve sensitive information. After the page data is collected, \textsc{Graph2Eval} utilizes LLMs to evaluate links and pages for privacy and compliance. The evaluation dimensions include: \textbf{(i)~privacy\_safe}: no exposure of personal or private information; \textbf{(ii)~copyright\_safe}: no proprietary or subscription content; \textbf{(iii)~content\_safe}: no harmful or inappropriate content; \textbf{(iv)~robots\_compliant}: compliance with \texttt{robots.txt} rules. Any page that fails the \textbf{privacy\_safe} check is automatically filtered out and does not proceed to subsequent graph construction or task generation processes.

\paragraph{Task Generation Stage.}
In the task generation stage, constraints are explicitly defined in the prompt templates for web tasks, prohibiting tasks involving payment or privacy data operations. These constraints are applied both in LLM prompts based on subgraph analysis and those based on meta-path analysis, ensuring that generated web tasks do not require agents to perform operations involving sensitive information. Additionally, the system employs a business data placeholder mechanism, using the placeholder \texttt{[BUSINESS\_DATA]} in task descriptions and steps to replace real data values. This prevents the generated tasks from directly containing any actual business data that might leak sensitive information. Task examples clearly demonstrate the use of placeholders instead of real data.

\section{Task Templates}
\label{sec:task_template}
Task templates constitute the core structural modules in \textsc{Graph2Eval} for automatically generating evaluation tasks. Each template is designed as a structured data class and encapsulates fundamental information including a \textit{template ID}, \textit{name}, \textit{description}, \textit{task type} and \textit{difficulty level}. In addition, templates provide Jinja2-formatted prompt and reference answer templates to dynamically generate task content.

Each template imposes strict \textbf{graph-structure requirements}, specifying mandatory node types, edge types, minimum and maximum node counts, and maximum hop distances, ensuring that tasks are instantiated from specific subgraph structures within the knowledge graph. \textsc{Graph2Eval} supports twelve distinct \textbf{text-based task types}, ranging from fundamental tasks such as information extraction, comprehension, and summarization, to more complex tasks including multi-hop reasoning, comparative analysis, fact verification, image interpretation, and cross-referencing. Each task type is associated with a designated \textbf{difficulty level} (Easy, Medium, Hard).

Templates also define detailed \textbf{evaluation criteria}, including evaluation metrics, requirements for exact matching, reference citations, reasoning paths, as well as version control and tagging mechanisms. The \textbf{template library manager} intelligently selects suitable templates based on the structural features of a given graph (node types, edge types, and node counts) and leverages the Jinja2 engine to render variables into concrete task prompts and reference answers. This facilitates fully automated instantiation from abstract templates to concrete evaluation tasks.

\begin{tcolorbox}[enhanced,
    breakable,
    colback=blue!5,
    colframe=blue!80!black,
    coltitle=white,
    fonttitle=\bfseries,
    title=Comparison Tasks in Task Templates,
    boxrule=0.8pt,
    arc=4pt,
    left=4mm, right=4mm, top=2mm, bottom=2mm,
    drop shadow=black!30!white]
Compare the following pieces of information:

\begin{itemize}
  \item Item 1: \texttt{\{\{ comparison\_items[0].content \}\}}
  \item Item 2: \texttt{\{\{ comparison\_items[1].content \}\}}
  % 可根据需要继续列出更多项
\end{itemize}

\texttt{\{\{ question \}\}}

Provide a detailed comparison and cite your sources.

\texttt{\{\{ answer \}\}}
\end{tcolorbox}

\section{Meta-Path}
\label{sec:meta_path}
The meta-path serves as a core mechanism in \textsc{Graph2Eval} for generating web-based tasks, enabling automatic transformation from subgraphs of web pages into executable tasks. The system employs a hierarchical design comprising two key components: \textit{MetapathPattern} and \textit{MetapathInstance}. The pattern defines the structural template of a task, $e.g.$, $
\text{SearchBox}(\$search) -[\text{Fills}]-> \text{BusinessData}(\$query) -[\text{Controls}]-> \text{Button}(\$submit)
$, while the instance represents the matching result of a pattern on a concrete subgraph. The system integrates a \textit{Graph Regex Engine}, supporting regex-like graph pattern syntax, including node type matching, edge type matching, quantifiers (e.g., ?, *, +, \{n,m\}), and alternative constructs (e.g., \texttt{Toast|Modal}), thereby enabling flexible matching and dynamic composition.

\textsc{Graph2Eval} follows a three-tier priority strategy: (1)~\textbf{Business Data Patterns}: require subgraphs to contain real business data nodes (e.g., user data, product data, order data), enabling generation of high-value business-related tasks; (2)~\textbf{General Interaction Patterns}: applicable to pages with common web elements such as search boxes, buttons, or navigation elements; (3)~\textbf{Basic Interaction Patterns}: serve as fallback mechanisms to ensure that even the simplest page structures can generate basic interactive tasks. Variables in patterns (e.g., \$search, \$button) are bound to specific node IDs in the subgraph via a slot-binding mechanism. Based on the matched pattern type, the system generates corresponding task steps (e.g., click, input, navigate), ultimately producing a fully executable web task instance. This framework achieves automated transformation from abstract graph structures to concrete, actionable tasks.

\section{Limitations and Discussion}
\label{appendix:limitations}
Our approach has two main limitations. (1) It relies on accurate knowledge graph modeling. If the graph is incomplete or contains errors, the quality of the generated tasks may be adversely affected. (2) The method currently focuses on structured knowledge and predefined relationships. This may limit its ability to generate tasks involving unstructured or novel information outside the knowledge graph, potentially constraining the diversity and coverage of tasks. In future work, these limitations could be addressed by integrating more robust knowledge-graph construction techniques and incorporating methods that allow the model to leverage unstructured data, thereby improving task quality, diversity, and coverage.

\section{Agent Architecture}
\label{sec:agent_design}

In this section, we describe the types of agents supported in \textsc{Graph2Eval}. Their capabilities are summarized in Table~\ref{tab:agent_capabilities}.

\paragraph{Single-Agent.} 
The Single-Agent integrates a Retrieval-Augmented Generation (RAG) framework, following a four-step \textit{retrieve-execute-evaluate-respond} workflow. It produces structured outputs that include references, reasoning paths, and confidence scores. A memory management module records task history and supports interactive dialogue, enabling more coherent and context-aware reasoning.

\paragraph{Multi-Agent.} 
The Multi-Agent establishes a distributed collaborative reasoning architecture for complex task decomposition and coordination, which also integrates RAG capabilities. It defines five core agent roles: \textit{PLANNER} (task planning and decomposition), \textit{RETRIEVER} (information retrieval), \textit{REASONER} (logical inference), \textit{VERIFIER} (result validation), and \textit{SUMMARIZER} (information consolidation). Each agent maintains independent reasoning capabilities, recording reasoning steps via \texttt{ReasoningStep} structures that capture source/target nodes, edge relations, logic, and confidence. Agents communicate through a standardized messaging protocol, enabling dynamic task allocation and load balancing. This design allows the system to handle complex multi-hop reasoning tasks, achieving collaborative reasoning performance superior to single-agent setups.

\paragraph{SoM Agent.} 
The SoM Agent integrates the SoM annotation to achieve precise element localization on web pages. Interactive elements are annotated with color-coded borders (e.g., M1, M2, M3), where each mark uniquely represents a specific element type. To locate a target element, the system generates a screenshot containing all marks and leverages a dedicated SoM analysis prompt to guide the LLM in identifying the corresponding mark ID, establishing a complete mapping from textual description to visual mark to exact coordinates.

\paragraph{Agent S 2.5.} 
Agent S 2.5 implements a reflective multimodal reasoning system. Its four-layer architecture comprises \texttt{LMMAgent} (multi-modal language model agent), \texttt{WebACI} (browser interface), \texttt{Worker} (execution agent with procedural memory and reflection), and \texttt{ProceduralMemory} (structured task guidance). The core ability lies in the reflection mechanism, where an independent module analyzes execution trajectories, identifies issues, and provides improvement suggestions. Multimodal input processing enables simultaneous analysis of screenshots and text, facilitating more accurate page understanding and element localization.

\begin{table*}[h!]
\centering
\setlength{\tabcolsep}{3.6pt}
\begin{tabular}{lcccc}
\toprule
\textbf{Capability} & \textbf{Single-Agent} & \textbf{Multi-Agent} & \textbf{SoM Agent} & \textbf{Agent S 2.5} \\
\midrule
\rowcolor{CadetBlue!20} \multicolumn{5}{c}{\underline{\textit{Architecture \& Knowledge}}} \\
\addlinespace[0.7mm]
Knowledge Retrieval & $\checkmark$ & $\checkmark$ & $\times$ & $\times$ \\
Memory Management & $\checkmark$ & $\checkmark$ & $\triangle$ & $\checkmark$ \\
\midrule
\rowcolor{CadetBlue!20} \multicolumn{5}{c}{\underline{\textit{Interaction \& Web Automation}}} \\
\addlinespace[0.7mm]
Web Automation & $\times$ & $\times$ & $\checkmark$ & $\checkmark$ \\
Visual Marking & $\times$ & $\times$ & $\checkmark$ & $\checkmark$ \\
Multimodal Processing & $\checkmark$ & $\checkmark$ & $\checkmark$ & $\checkmark$ \\
\midrule
\rowcolor{CadetBlue!20} \multicolumn{5}{c}{\underline{\textit{Reasoning \& Evaluation}}} \\
\addlinespace[0.7mm]
Task Planning & $\triangle$ & $\checkmark$ & $\checkmark$ & $\checkmark$ \\
Error Handling & $\triangle$ & $\checkmark$ & $\checkmark$ & $\checkmark$ \\
Reflection  & $\times$ & $\times$ & $\times$ & $\checkmark$ \\
\midrule
\rowcolor{CadetBlue!20} \multicolumn{5}{c}{\underline{\textit{Execution \& Performance}}} \\
\addlinespace[0.7mm]
Concurrency & $\times$ & $\triangle$ & $\times$ & $\times$ \\
Token Monitoring & $\checkmark$ & $\checkmark$ & $\checkmark$ & $\checkmark$ \\
\bottomrule
\end{tabular}
\caption{Capability Comparison of the Four Agents in the \textsc{Graph2Eval} Framework. $\checkmark$ indicates a fully supported capability, 
$\triangle$ indicates partial support, 
and $\times$ indicates that the capability is not supported.}
\label{tab:agent_capabilities}
\end{table*}

\section{Metrics Formula}
\label{sec:metrics_formula}

\subsection{Document Understanding Task Metrics}

\paragraph{F1 Score.} 
For a predicted answer span $P$ and a ground-truth span $G$, precision and recall are defined as
\begin{equation}
\text{Precision} = \frac{|P \cap G|}{|P|}, \quad
\text{Recall} = \frac{|P \cap G|}{|G|}.
\end{equation}
The F1 score is their harmonic mean:
\begin{equation}
\text{F1} = \frac{2 \cdot \text{Precision} \cdot \text{Recall}}
{\text{Precision} + \text{Recall}}.
\end{equation}

\paragraph{ROUGE-L.} 
ROUGE-L measures the longest common subsequence (LCS) between $P$ and $G$:
\begin{equation}
\text{ROUGE-L} = \frac{(1 + \beta^2) \cdot \text{R}_{\text{LCS}} \cdot \text{P}_{\text{LCS}}}
{\text{R}_{\text{LCS}} + \beta^2 \cdot \text{P}_{\text{LCS}}},
\end{equation}
where
\[
\text{R}_{\text{LCS}} = \frac{\text{LCS}(P,G)}{|G|}, \quad
\text{P}_{\text{LCS}} = \frac{\text{LCS}(P,G)}{|P|}.
\]

Here, \(\beta\) is a weighting parameter that balances the importance of recall versus precision. 
Following common practice, we set \(\beta = 1\) to weigh precision and recall equally.

\paragraph{LLM-as-a-Judge.} 
LLM is prompted to evaluate the predicted answer $P$ against the ground-truth $G$ 
and task instruction along multiple dimensions, including quality, relevance, and completeness. 
The scores are normalized to $[0,1]$, and can be aggregated into an overall similarity judgment, 
with higher values indicating stronger semantic alignment. The instruction used to guide the LLM in this evaluation are provided below.

\begin{tcolorbox}[
    enhanced,
    breakable,
    colback=blue!5,         % 背景色
    colframe=blue!80!black, % 边框色
    coltitle=white,
    fonttitle=\bfseries,
    title=LLM Instructions for Evaluating Document Understanding Tasks,
    boxrule=0.8pt,          % 边框粗细
    arc=4pt,                % 圆角
    left=4mm, right=4mm, top=2mm, bottom=2mm, % 内边距
    drop shadow=black!30!white
]
You are an expert evaluator assessing the quality of an AI-generated answer. Please evaluate the following:

TASK: \{task.prompt\} \\
GOLD STANDARD ANSWER: \{gold\_answer\} \\
GENERATED ANSWER: \{pred\_answer\}

Rate the generated answer on these 3 key dimensions (0.0 to 1.0):

1. ANSWER\_QUALITY: Overall quality and accuracy of the answer compared to the gold standard \\
2. RELEVANCE: How well the answer addresses the specific task/question \\
3. COMPLETENESS: How complete and comprehensive the answer is

Provide your assessment in JSON format:
\begin{verbatim}
{
    "answer_quality": <score>,
    "relevance": <score>,
    "completeness": <score>
}
\end{verbatim}

Be objective and focus on the most important aspects of answer quality.
\end{tcolorbox}

\subsection{Web Task Metrics}

\paragraph{Success Rate (SR).} 
We evaluate web interaction tasks using \textit{Success Rate (SR)}, defined as the fraction of tasks judged successful by an LLM evaluator:
\begin{equation}
\text{SR} = \frac{N_{\text{success}}}{N_{\text{total}}},
\end{equation}
where $N_{\text{success}}$ is the number of tasks determined to be successfully completed by LLMs, 
and $N_{\text{total}}$ is the total number of tasks. The use of LLMs for evaluation is motivated by the fact that web interaction tasks are executed in live, dynamic online environments, 
where the complexity and variability of web pages render rule-based evaluation (e.g., checking system states or explicit signals) unreliable. 
To address this challenge, we employ LLMs to determine task success by analyzing the sequence of executed actions, 
the final page state, and any encountered error messages. 
By leveraging the LLM’s ability to interpret online content and reason about task goals, 
this approach provides a consistent, scalable, and generalizable measure of task completion. 
The instructions used to guide the LLM in evaluating task success are provided below.

\subsection{LLM Evaluator Reliability}
\label{appendix:LLM_Evaluator}
Using LLMs as evaluators to assess task success rate in dynamic Web environments offers high scalability but may raise reliability concerns. To validate the reliability of LLM-as-a-Judge on web interaction tasks, We randomly sampled 50 task trajectories completed by Agent S 2.5 from 317 web interaction tasks. Two human annotators independently judged whether each trajectory successfully completed the task (“success” or “failure”), with agreement indicating success. The results show that 88\% of the judgments matched those of \texttt{GPT-4o}, demonstrating high consistency.

\begin{figure*}
\begin{tcolorbox}[
    enhanced,
    breakable,
    colback=blue!5,
    colframe=blue!80!black,
    coltitle=white,
    fonttitle=\bfseries,
    title=LLM Instructions for Evaluating Web Interaction Tasks,
    boxrule=0.8pt,
    arc=4pt,
    left=4mm, right=4mm, top=2mm, bottom=2mm,
    drop shadow=black!30!white
]
Task: \{task.prompt if hasattr(task, 'prompt') else f'Complete task: \{task.task\_id\}'\}

Execution Summary: \\
- Actions executed: \{len(trajectory.actions\_executed)\} \\
- Success: \{trajectory.success\} \\
- Error message: \{trajectory.error\_message or 'None'\} \\

Current page URL: \{page\_info.get('url', 'Unknown')\} \\
Current page title: \{page\_info.get('title', 'Unknown')\} \\

Actions executed: \\
\{chr(10).join([f"- \{action.get('action', 'unknown')\}" for action in trajectory.actions\_executed])\} \\

Please evaluate if the task has been completed successfully by analyzing the current page state. Consider:  
1. Whether all required actions were performed  
2. Whether the final state matches the task requirements  
3. Whether any errors occurred that prevent completion  
4. Whether the current page content indicates task completion  

Respond with valid JSON format (no markdown, no code blocks):
\begin{verbatim}
{
    "task_completed": true,
    "confidence": 0.8,
    "reasoning": "explanation of your evaluation",
    "missing_actions": ["list of any missing actions"],
    "final_state_analysis": "description of current page state"
}
\end{verbatim}
\end{tcolorbox}
\end{figure*}

\section{\textsc{Graph2Eval-Bench} Overview}
\label{sec:dataset_overview}
\begin{figure*}[h]
    \centering
    % --------------------------
    % 左侧表格
    % --------------------------
    \begin{minipage}[b]{0.5\textwidth} % 宽度可调
        \centering
        \small
        \setlength{\tabcolsep}{3.4pt}
        \renewcommand{\arraystretch}{1.2} % 行高
        \begin{tabular}{l c}
            \toprule
            \textbf{Data Source} & \textbf{Tasks Num.} \\
            \midrule
            \rowcolor{CadetBlue!20} \multicolumn{2}{c}{\underline{\textbf{Document Understanding Datasets}}} \\
            \addlinespace[0.7mm]
            Agent AI & 64 \\
            AgentHarm & 65 \\
            AI Agent under Threat & 50 \\
            The Dawn of GUI Agent & 68 \\
            Data Shapley & 63 \\
            DeepSeek-R1 & 58 \\
            Speculative Decoding & 70 \\
            GPT-4o System Card & 72 \\
            learning\_dynamic & 67 \\
            lightRAG & 68 \\
            Navigating the Risks & 47 \\
            OpenAI o3 and o4-mini & 62 \\
            OS Agents & 64 \\
            Qwen-VL & 58 \\
            RTBAS & 61 \\
            TaskCraft & 59 \\
            \midrule
            \rowcolor{CadetBlue!20} \multicolumn{2}{c}{\underline{\textbf{Web Interaction Datasets}}} \\
            \addlinespace[0.7mm]
            Mozilla Developer & 28 \\
            GitHub & 17 \\
            Project Gutenberg & 15 \\
            Open Library & 78 \\
            OpenWeather & 16 \\
            Stack Overflow & 29 \\
            The Guardian & 87 \\
            WIRED & 47 \\
            \bottomrule
        \end{tabular}
        \captionof{table}{Task distribution across different data sources in \textsc{Graph2Eval-Bench}.}
        \label{tab:datasets}
    \end{minipage}%
    \hfill
    % --------------------------
    % 右侧图表
    % --------------------------
    \begin{minipage}[b]{0.48\textwidth} % 宽度可调
        \centering
        \includegraphics[width=\textwidth]{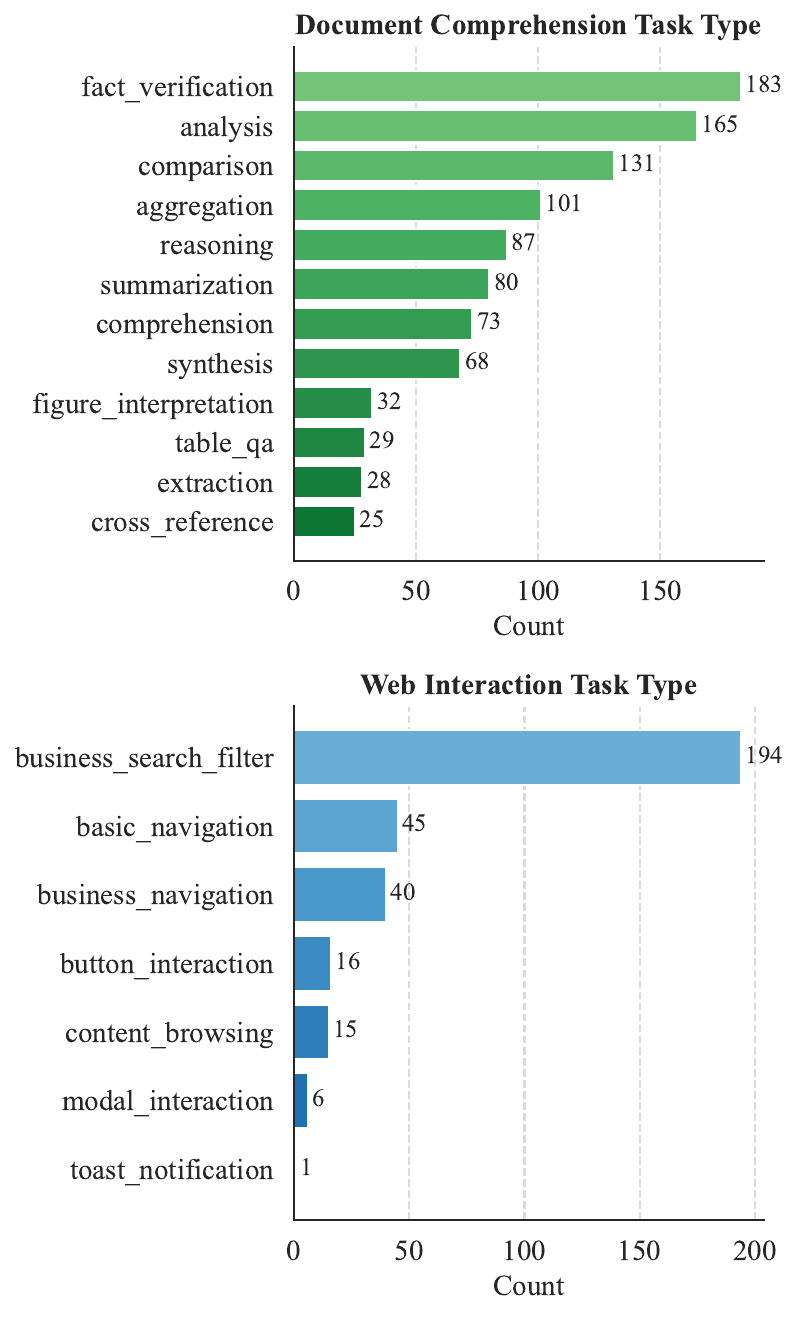}
        \caption{The number of tasks for each task type in \textsc{Graph2Eval-Bench}.}
        \label{fig:task_counts}
    \end{minipage}
\end{figure*}

Table \ref{tab:datasets} and Figure \ref{fig:task_counts} summarize the data sources of the \textsc{Graph2Eval-Bench}, the number of tasks generated from each source, and the distribution across task types. In total, the \textsc{Graph2Eval-Bench} comprises 1,319 tasks and consists of two primary components: document understanding datasets and web interaction datasets. The document understanding datasets cover a wide range of task types, including reasoning, analysis, and aggregation, and enable diverse evaluations across multiple modalities. To mitigate potential risks to live websites, the web interaction datasets focus on navigation and interaction tasks, spanning various domains such as digital libraries, weather services, and news portals.

\begin{figure*}[t]
    \centering
    \includegraphics[width=0.85\textwidth]{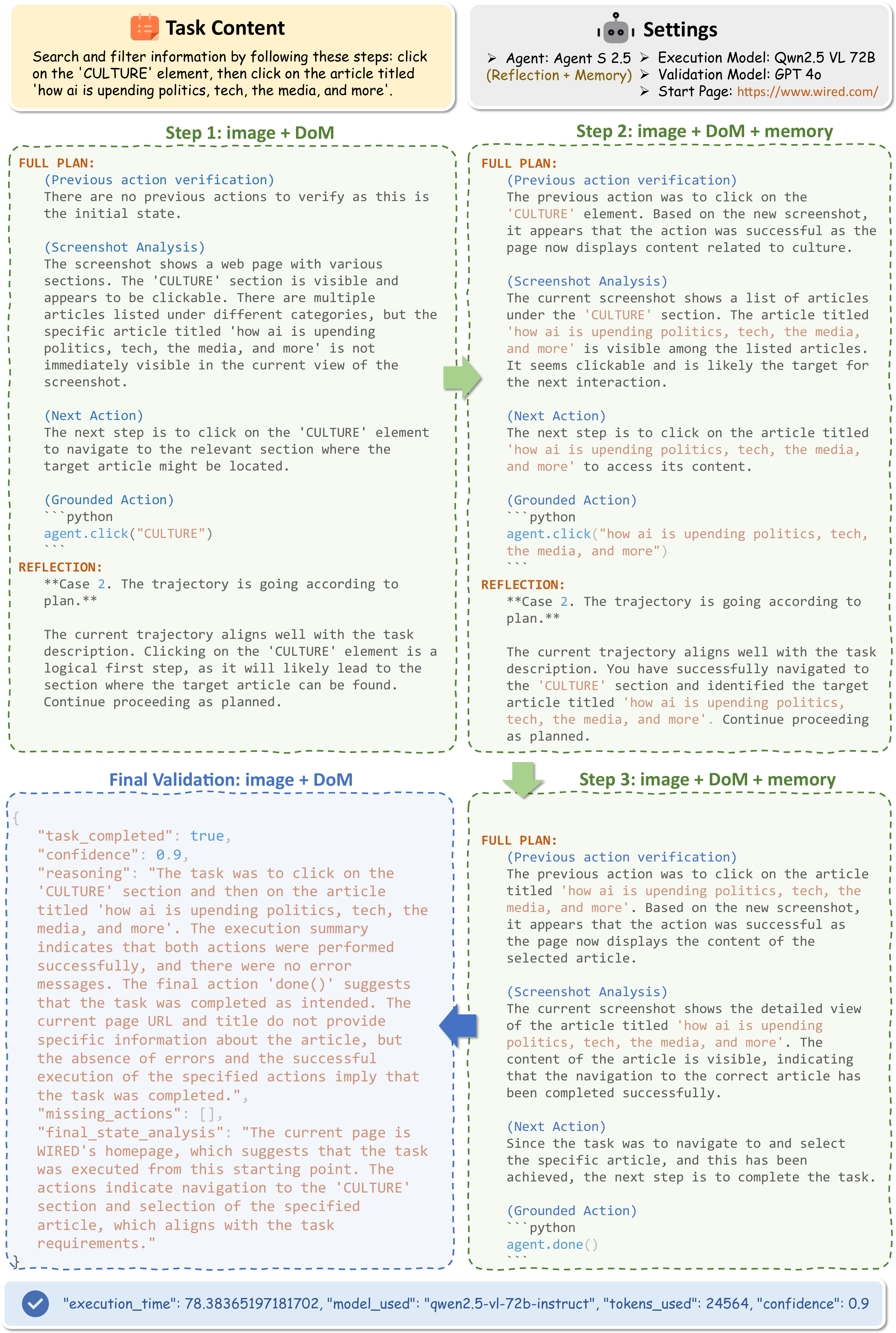}
    \caption{Case study of Agent S performing tasks on the Web dataset.}
    \label{fig:case_study}
\end{figure*}

\section{Case Study}
\label{sec:case_study}
We illustrate in Figure \ref{fig:case_study} the overall workflow of Agent S in performing a web interaction task. The figure presents both the task specification and the agent settings. Given page screenshots and DOM representations, Agent S 2.5 leverages its memory and reflection mechanisms to reason about the current state, determine the next action, and execute it accordingly. The final outcome of the task is subsequently validated by another LLM, ensuring the reliability of the result. During this process, \textsc{Graph2Eval} continuously monitors and records key performance indicators, including execution time and token consumption, which provide quantitative insights into the efficiency and cost of the agent. This case not only demonstrates how Agent S operates in a realistic web environment but also highlights the effectiveness of combining reasoning, memory, and external validation for reliable web-based task execution.

\section{Safety Task Generation}
\label{sec:safety_task_generation}

We investigate the generation of safety-focused tasks with the aim of improving the evaluation of agents’ safety boundaries. Building on the overall \textsc{Graph2Eval} framework, we develop an initial pipeline for synthesizing safety tasks. For text-understanding benchmarks, we derive safety-oriented instances from existing tasks so as to preserve task naturalness while minimally perturbing inputs, thereby amplifying the ability to detect latent model risks.

\paragraph{Safety Task Generation for Document Understanding.}

The generation pipeline is organized into three stages: \textbf{(1)~Policy-document parsing and threat extraction}: Given policy or security documents as input, LLMs are used to automatically extract candidate threat types and convert them into structured representations (e.g., threat category, examples, keywords, and severity) for downstream use. \textbf{(2)~Threat-embedding strategy exploration}: We explore multiple embedding strategies---content injection, prompt manipulation, context switching, and indirect reference---to integrate threat information into original texts in natural, covert, or semantically implicit forms. This enables assessment of models’ ability to recognize and defend against explicit and implicit threats under different surface realizations. \textbf{(3)~Safety-instance creation and preliminary quality control}: For each embedding strategy, we generate task instances and log metadata (task ID, difficulty level, prompt and reference answer, requisite reasoning traces, and evaluation criteria). We further perform an initial quality assessment of clarity, relevance, difficulty, and completeness, retaining only samples that meet preset thresholds to ensure test reliability.

\paragraph{Safety Task Generation for Web Interactions.}

For web interaction tasks, we primarily explore threat modeling in web environments. Because directly embedding threats into live online environments would create unacceptable real-world risks, we adopt a controlled sandbox approach: in isolated Docker environments or controlled browser sessions, we inject malicious environment artifacts via scripts. (e.g., forged phishing elements, malicious forms, or suspicious redirects) to simulate online threat scenarios safely and reproducibly. We further augment existing web interaction tasks with safety nudges (e.g., security warnings) and generate web-oriented safety tasks to evaluate models’ security judgment in web environments.


\end{document}